\documentclass[10pt,twocolumn,letterpaper]{article}

\usepackage[final,pagenumbers]{cvpr} % To force page numbers, e.g. for an arXiv version
\usepackage[utf8]{inputenc} % allow utf-8 input

\DeclareUnicodeCharacter{2662}{$\diamondsuit$} % for ♢
\DeclareUnicodeCharacter{2661}{$\heartsuit$}    % for ♡
\DeclareUnicodeCharacter{2217}{$\ast$}          % for ∗

\usepackage[T1]{fontenc}    % use 8-bit T1 fonts
\usepackage{hyperref}       % hyperlinks
\usepackage{url}            % simple URL typesetting
\usepackage{booktabs}       % professional-quality tables
\usepackage{amsfonts}       % blackboard math symbols
\usepackage{nicefrac}       % compact symbols for 1/2, etc.
\usepackage{microtype}      % microtypography
\usepackage{graphicx}
\usepackage{caption}
\usepackage{subcaption}
\usepackage{float}
\usepackage{algorithm}
\usepackage{algpseudocode}
\usepackage{amsmath}
\usepackage{array}
\usepackage{tabularx}  % For flexible table columns\
\usepackage{textcomp}
\definecolor{cvprblue}{rgb}{0.21,0.49,0.74}
\usepackage{xcolor, colortbl}

\definecolor{darkred}{rgb}{0.7,0.1,0.1}
\definecolor{darkgreen}{rgb}{0.1,0.6,0.1}
\definecolor{cyan}{rgb}{0.7,0.0,0.7}
\definecolor{otherblue}{rgb}{0.1,0.4,0.8}
\definecolor{maroon}{rgb}{0.76,.13,.28}
\definecolor{burntorange}{rgb}{0.81,.33,0}

\usepackage{pifont}
\newcommand{\cmark}{\color{darkgreen}{\ding{51}}}
\newcommand{\xmark}{\color{darkred}{\ding{55}}}

\def\paperID{16847} % *** Enter the Paper ID here
\def\confName{CVPR}
\def\confYear{2025}

\title{TextInVision: Text and Prompt Complexity Driven Visual Text Generation Benchmark}

\author{%
  Forouzan Fallah$^{\star}$, Maitreya Patel$^{\star}$, Agneet Chatterjee$^{\star}$, Vlad I. Morariu$^{\lozenge}$, Chitta Baral$^{\star}$, Yezhou Yang$^{\star}$\\[1ex]
  $^{\star}$Arizona State University \quad $^{\lozenge}$Adobe Research
}

\begin{document}
\maketitle
\begin{abstract}

Generating images with embedded text is crucial for the automatic production of visual and multimodal documents, such as educational materials and advertisements. However, existing diffusion-based text-to-image models often struggle to accurately embed text within images, facing challenges in spelling accuracy, contextual relevance, and visual coherence. Evaluating the ability of such models to embed text within a generated image is complicated due to the lack of comprehensive benchmarks. In this work, we introduce TextInVision, a large-scale, text and prompt complexity driven benchmark designed to evaluate the ability of diffusion models to effectively integrate visual text into images. We crafted a diverse set of prompts and texts that consider various attributes and text characteristics. Additionally, we prepared an image dataset to test Variational Autoencoder (VAE) models across different character representations, highlighting that VAE architectures can also pose challenges in text generation within diffusion frameworks. Through extensive analysis of multiple models, we identify common errors and highlight issues such as spelling inaccuracies and contextual mismatches. By pinpointing the failure points across different prompts and texts, our research lays the foundation for future advancements in AI-generated multimodal content.
Our dataset and benchmark can be found \href{https://github.com/TextinVision/TextinVision}{here}.

\end{abstract}

\section{Introduction}
\label{sec:intro}

Denoising diffusion probabilistic models \cite{dhariwal2021diffusion, ramesh2022hierarchical, saharia2022photorealistic, song2020denoising} have significantly boosted the development of general text-to-image (T2I) generation, demonstrating the capability of generating surprisingly high-quality images over the past few years. Despite their impressive achievements, these models exhibit a notable limitation: difficulty in generating images that include specific visual text. Image generation with embedded text is a critical task with wide-ranging applications, from advertising, where the precise rendering of brand names on products can alter consumer perception, to educational resources, where accurate depictions of text in diagrams or illustrations can significantly impact learning outcomes. However, the State-Of-The-Art (SOTA) diffusion models often struggle with this task \cite{rombach2022high}. 

\begin{figure}[]
    \centering
        \includegraphics[width=.49\textwidth]{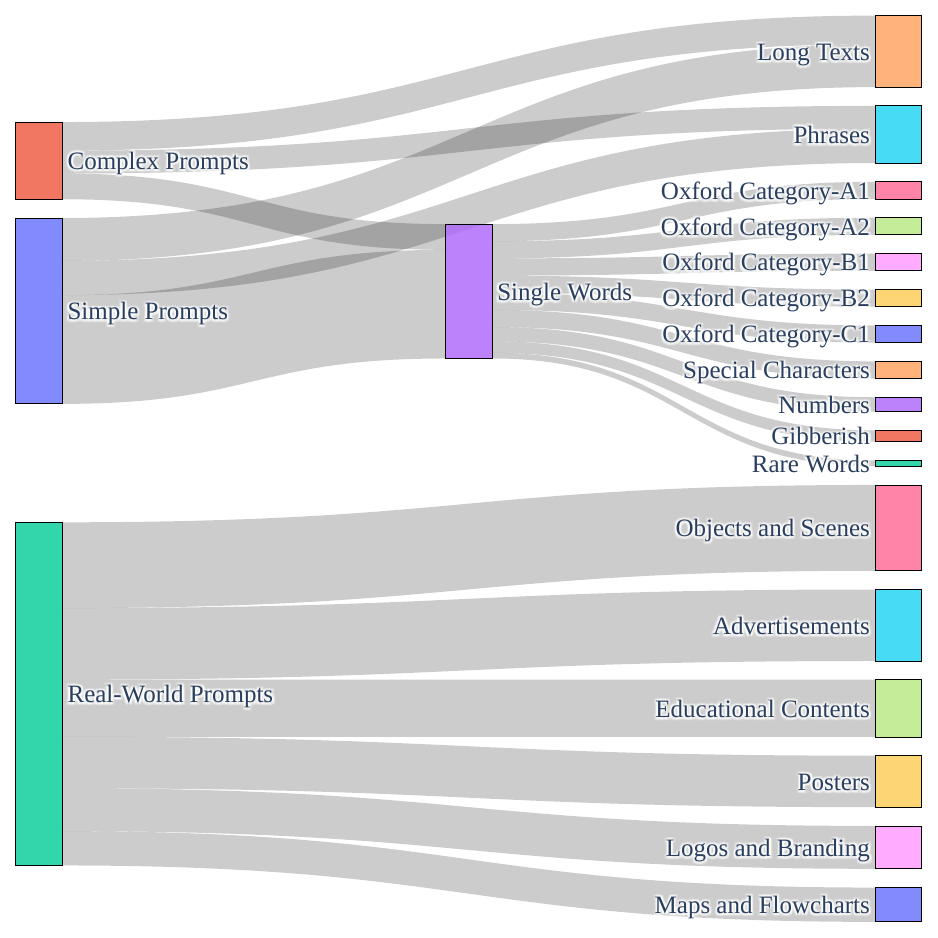}
        \caption{A comprehensive benchmark that enables separate analysis of prompt complexity and text attributes in evaluating T2I models. The benchmark features diverse prompt types—including Simple prompts (e.g., "a paper with the word 'text' written on it"), Complex prompts (e.g., "A sunny day at a theme park, a roller coaster with cars ascending the track, each car labeled with the text spelling out 'text', excited riders waving their hands"). It also incorporates a range of text attributes from single words to gibberish, numbers, special characters, phrases, and long texts.}

    \label{fig:bench_dist}
    \vspace{-0.5cm}
\end{figure}

\begin{figure*}[]
    \centering
        \begin{subfigure}[b]{\textwidth}
        \centering
        \includegraphics[width=\textwidth]{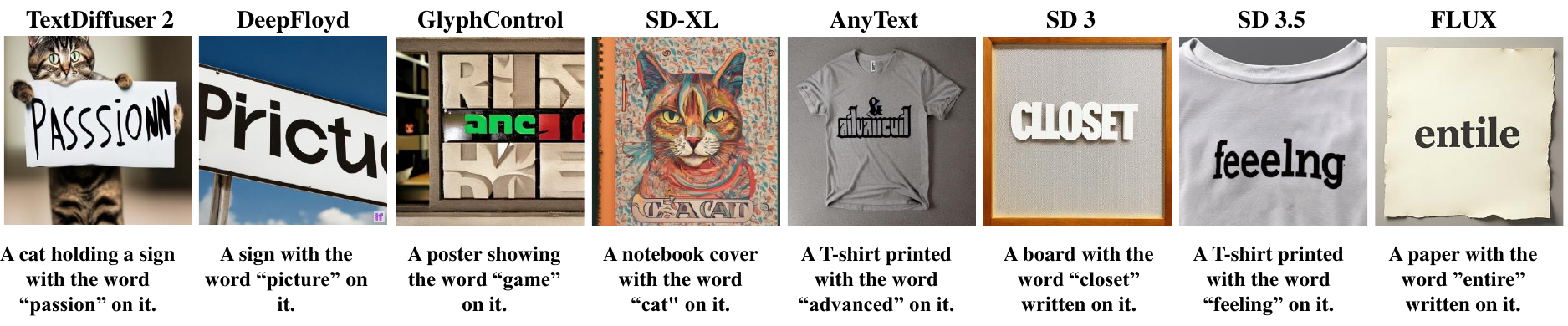}
        \caption{Sample images generated using simple prompts and single words.}
    \end{subfigure}
    \hfill
    \begin{subfigure}[b]{\textwidth}
        \centering
        \includegraphics[width=\textwidth]{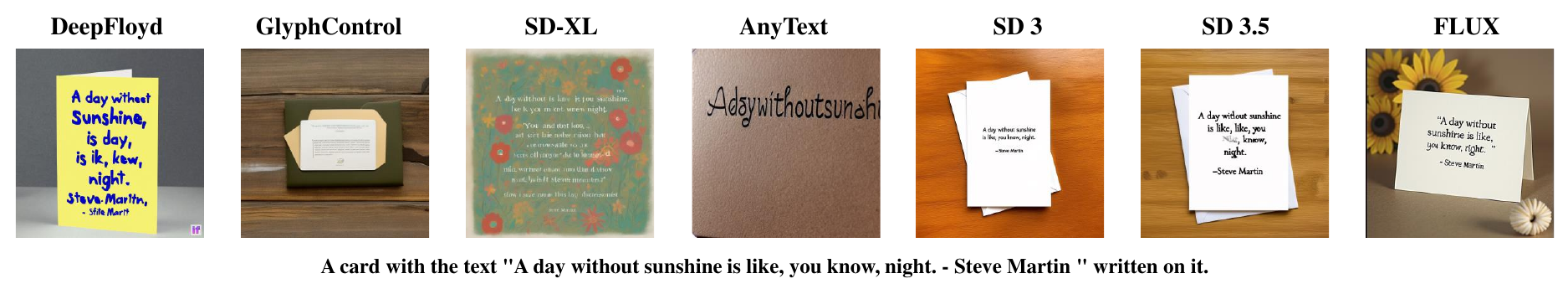}
        \caption{Sample images generated using simple prompts and long texts.}
    \end{subfigure}
    \hfill
    \begin{subfigure}[b]{\textwidth}
        \centering
        \includegraphics[width=\textwidth]{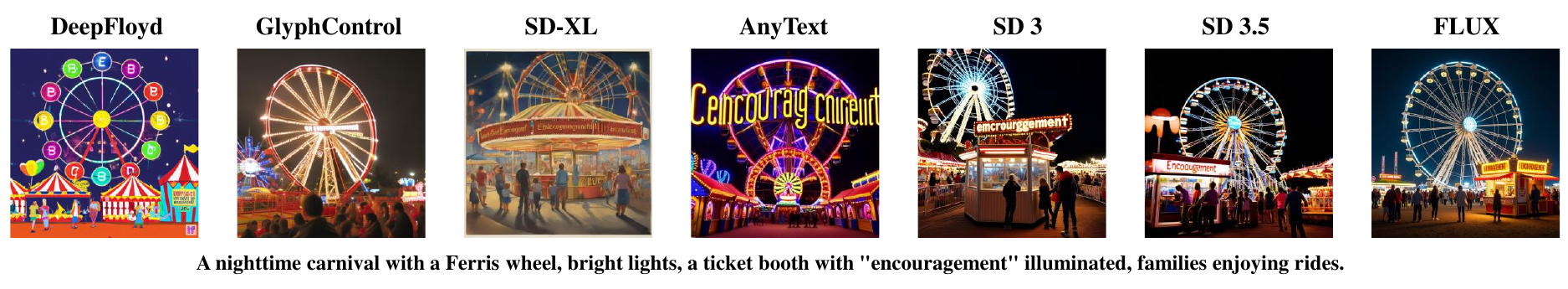}
        \caption{Sample images generated using complex prompts and single words.}
    \end{subfigure}
    \caption{Generated images from multiple T2I models demonstrating common limitations in accuracy, clarity, and legibility.}
    \label{fig:models_image}
    \vspace{-0.4cm}
\end{figure*}

Several research efforts have attempted to address visual text rendering in image generation \cite{GlyphControl, tuo2023anytext, CharacterAware, Textdiffuser,chen2023textdiffuser, balaji2022ediff}.
Although these models significantly improve the fidelity of generated visual text, they generally fall short in rendering longer textual elements\cite{betker2023improving, wiles2024revisiting, gani2023llm}. Figure \ref{fig:models_image} illustrates the challenges of accurate text rendering in different models. This limitation not only restricts the utility of these models but also highlights a gap in our understanding of how to effectively integrate textual and visual information in generated content. Moreover, existing models tend to default to generating common words or phrases, substituting user-specified unique terms, names, or phrases---often unique or novel---in the generated images. This behavior detracts from the personalized experience users seek and limits the applicability of these models in domains requiring high specificity and accuracy. 

A significant obstacle in advancing this area is the absence of comprehensive benchmarks specifically designed to evaluate the ability of T2I models to generate images containing accurate and contextually relevant text. In this paper, we address this gap by introducing TextInVision, a large-scale benchmark tailored to evaluate the integration of visual text in image generation models. Our key contributions are as follows:

\begin{enumerate}

    \item We present TextInVision, the first-of-its-kind visual text generation benchmark that mirrors real-world use cases, driven by word and prompt complexity. This benchmark emphasizes practical applications and challenges encountered outside controlled experimental settings, establishing a new standard for evaluating image generative models in dynamic environments.

    \item We craft a diverse set of texts and prompts that consider various attributes and characteristics. Our methodology involves a carefully curated selection, systematically categorized to ensure a thorough examination of models' performance across a wide range of textual inputs. By independently varying prompt complexity and text attributes, this benchmark provides a detailed assessment of how well T2I models embed readable and meaningful text within different visual contexts.

    \item We prepare an image dataset designed to test Variational Autoencoder (VAE) models across different character representations. This highlights challenges posed by VAE architectures in text generation within diffusion frameworks, an area less explored in existing literature.

    \item Through comprehensive experiments using multiple open-source models, we rigorously dissect the T2I translation process to identify the origins of errors and evaluate their impact. Our key findings are: a) Word frequency impacts performance less than previously assumed. b) The visual encoder is a major contributor to error. c) Rare words present substantial challenges to T2I models, with gibberish prompts unexpectedly outperforming rare words in preserving translation quality. d) Word length critically influences performance: models generate multiple short words more effectively than a single lengthy word.
\end{enumerate}

\section{Related work}
\label{sec:relatedwork}

\paragraph{Visual Text Generation Models}
Recent T2I diffusion models have significantly enhanced their text generation capabilities \cite{flux, sd3, gu2022vector, Denoising, rombach2022high, zhang2023adding, nichol2021glide}. In visual text generation, these models have been adapted to focus on generating more readable and visually clear text. 
 For instance, Stable Diffusion (SD) trained with data and a powerful CLIP text encoder are proficient at generating high-quality images but often fail to generate precise and accurate visual text \cite{Hierarchical}. 
Building upon SD, SDXL \cite{rombach2022high} improves Stable Diffusion by using a 3× larger U-Net and a second text encoder and introducing another refinement model to improve image quality. According to \cite{CharacterAware}, incorporating a Character-aware text encoder results in significantly fewer spelling errors compared to using Character-blind encoders. Training a character-aware text encoder and an inpainting version of Stable Diffusion, over extensive datasets,  UDiffText \cite{UDiffText} results in significant advancements. GlyphControl \cite{GlyphControl} improves the performance of SD models in generating accurate visual text by using additional glyph conditional information. In TextDiffuser \cite{Textdiffuser}, the process starts by extracting keywords from the text prompts to create a layout, which is then used to generate images by model from U-Net \cite{unet}. In TextDiffuser-2 \cite{chen2023textdiffuser}, a language model for layout planning improves spelling accuracy through the use of fine-grained tokenizers, such as character and position tokens from \cite{CharacterAware}. Diff-Text \cite{BrushYourText} introduces a training-free framework for generating multilingual visual text images. It utilizes attention constraints and image-level constraint within the U-Net’s cross-attention layer to improve the accuracy of visual text. AnyText \cite{tuo2023anytext} comprises a text-control diffusion pipeline with an auxiliary latent module and text embedding module, which can generate multilingual visual text. These developments demonstrate the ongoing efforts to improve visual text generation within diffusion models. However, effectively evaluating these models requires robust benchmarks that can accurately assess their capabilities.

\paragraph{Visual Text Generation Evaluation}
The evaluation of T2I models is a crucial step in understanding their effectiveness and identifying areas for improvement. To thoroughly evaluate T2I models, the choice of the prompts is key, as it determines which abilities or skills of the models are being evaluated \cite{wiles2024revisiting}. The MARIO-Eval \cite{Textdiffuser} is a benchmark designed to assess text rendering quality in images, including a collection of prompts sourced from the MARIO-10M test set and other datasets. The LAION-Glyph \cite{GlyphControl} introduces two evaluation benchmarks, SimpleBench and CreativeBench. Words for these prompts are selected from Wikipedia and categorized into four frequency-based buckets. Gecko \cite{wiles2024revisiting}, although useful, is limited because it is not designed specifically for the text rendering task and lacks a comprehensive set of prompts to thoroughly evaluate models. LenCom-EVAL \cite{RefiningTextImage} is designed to assess the robustness of T2I models by challenging them with prompts that include special characters, numbers, and lengthy text. However, its effectiveness is constrained by a maximum prompt length of 17 words, and it is not publicly available, which limits broader access and comparative analysis.

Despite recent efforts, current benchmarks for T2I models often rely on unstructured and arbitrary selections of prompts, failing to systematically evaluate models capability on visual text generation. To address this gap, we introduce TextInVision, a comprehensive benchmark that evaluates models by separately analyzing varying levels of prompt complexity and a wide range of text attributes. By including an image dataset specifically designed to evaluate the VAE components of T2I models, TextInVision enables a comprehensive assessment of the models for text fidelity.

\section{TextInVision benchmark}
\label{bench}
In this section, we introduce a novel benchmark for evaluating visual text generation, designed to evaluate their performance across varying levels of complexity from simple tasks to real-world scenarios. Our benchmark addresses practical applications and challenges beyond controlled settings, establishing a new standard for assessing these models. Table \ref{tab:benchmark} presents a comparison between TextInVision attributes and other existing benchmarks used to evaluate the visual text rendering task. To create meaningful evaluation prompts, we gathered diverse real-world scenarios such as advertisements, promotional materials, and educational content. Recognizing that current models often struggle with complex prompts and integration issues, we included both intricate and simpler prompts in our benchmark. This approach allows us to incrementally track model performance and identify specific areas for improvement in model design and fine-tuning. TextInVision comprises over 50,000 methodically designed prompts, assessing models' capabilities across a broad spectrum of scenarios. By independently varying prompt complexity and text attributes, our benchmark provides a detailed assessment of how well T2I models embed readable and meaningful text within different visual contexts. Appendix A contains further details and multiple examples.

\begin{table*}[t]
\centering
\caption{Comparison of benchmark characteristics for evaluating visual text rendering tasks.}
\resizebox{0.7\textwidth}{!}{%
\begin{tabular}{@{}lccccc>{\columncolor[gray]{0.15}}c@{}}
\toprule
Features & MARIO-Eval & GlyphControl & LenCom-Eval & Gecko & TextInVision (Ours) \\ \midrule
Num. Prompts & 5414 & 400+ & 3000 & 1500  & 50000+ \\
Visual Comp. & \xmark & \cmark & \cmark &  \cmark & \cmark\\
text style & \xmark & \cmark & \cmark &  \cmark & \cmark\\
Freq. COCO, LAION & \xmark & \xmark & \xmark  & \xmark & \cmark\\
Spec. Char. Num & \xmark & \xmark & \cmark & \cmark & \cmark\\
Gibberish Text & \xmark & \xmark & \cmark  & \cmark & \cmark\\
Text Comp. Categ. & \xmark & \xmark & \xmark  & \xmark & \cmark\\
Text Length Categ. & \xmark & \xmark & \cmark  & \xmark & \cmark\\
Public Avail. & \cmark & \cmark & \xmark  & \cmark  & \cmark\\ \bottomrule
\end{tabular}%
}
\label{tab:benchmark}
\end{table*}

\subsection {Prompt selection}
We employ a systematic approach to derive prompts from real-world scenarios, ensuring comprehensive testing across various text-in-image applications. Our selection process involves categorizing images into distinct scenarios, each reflecting typical contexts where text-inclusive images occur. Specifically, we define categories such as professional settings, commercial environments, and educational contexts. For instance, professional settings are represented by items like corporate reports and official documents, which assess the models' precision and formality in text presentation. Commercial signage, including store signs and safety notices, evaluates clarity and impact in retail and safety-critical environments. Educational materials test the models' ability to render informative text in diagrams and instructional images. In addition to these diverse and complex scenarios, we incorporate a selection of very simple prompts. By including simpler prompts, such as single-word labels or basic phrases, we evaluate the models' effectiveness in accurately rendering text with clarity and precision in less demanding contexts. This comprehensive approach ensures that we can pinpoint the models' strengths and weaknesses across different application domains.

\subsection{Text selection}
We choose different words and phrases from real images that contain text to ensures our prompts are realistic and relevant. We categorize the selected text into single words, phrases, and long text, with lengths varying from one word to several sentences. To assess the models' ability to handle text of varying complexity, we utilize the Oxford 5000 by CEFR level \cite{oxford5000}. By including words across all language proficiency levels, from basic (A1) to advanced (C1), we test the models' performance on text that ranges in difficulty and importance. In addition to text complexity, we consider word frequency by drawing data from the COCO \cite{coco} and LAION \cite{laion} datasets. By selecting words that occur at high, medium, and low frequencies, we evaluate whether the models exhibit any bias toward commonly seen words or can handle rare words with equal proficiency. This analysis helps us understand how the frequency of words used during training influences model performance.

To further challenge the models, we incorporate inputs beyond plain alphabetic characters. This includes gibberish text, intentional misspellings, special characters (e.g., $@$, \#, \$), numbers, and long rare words. By including such varied text elements, we test whether the models can accurately reproduce the provided text exactly as given, including any errors or unconventional characters. This aspect is crucial for assessing the models' ability to handle precise input, which is important for applications requiring high fidelity and accuracy.

Finally, we vary the length of the text, measured by the number of letters or words in phrases or sentences. This variability allows us to test the models’ performance as a function of different lengths. Short text demands precise placement and clarity in small spaces, while longer text challenges the models to maintain readability and aesthetic integration over larger areas. By ensuring that the models can effectively render text of any length, we address practical application needs where text length can vary significantly.

Having described the TextInVision benchmark, we transition to conducting experiments and analyses that leverage it to assess and comprehend the performance of T2I models.

\section{Experimental design and results}
\label{exp}

This section details the experiments and analysis performed with the TextInVision Benchmark. We report results on end-to-end image generation methodologies, explore the role of the VAE component in setting performance upper bounds for diffusion models and evaluate the performance of visual text image generation. 

\subsection{Real-world scenarios prompts}
\begin{figure}[t]
\centering
        \includegraphics[width=0.47\textwidth]{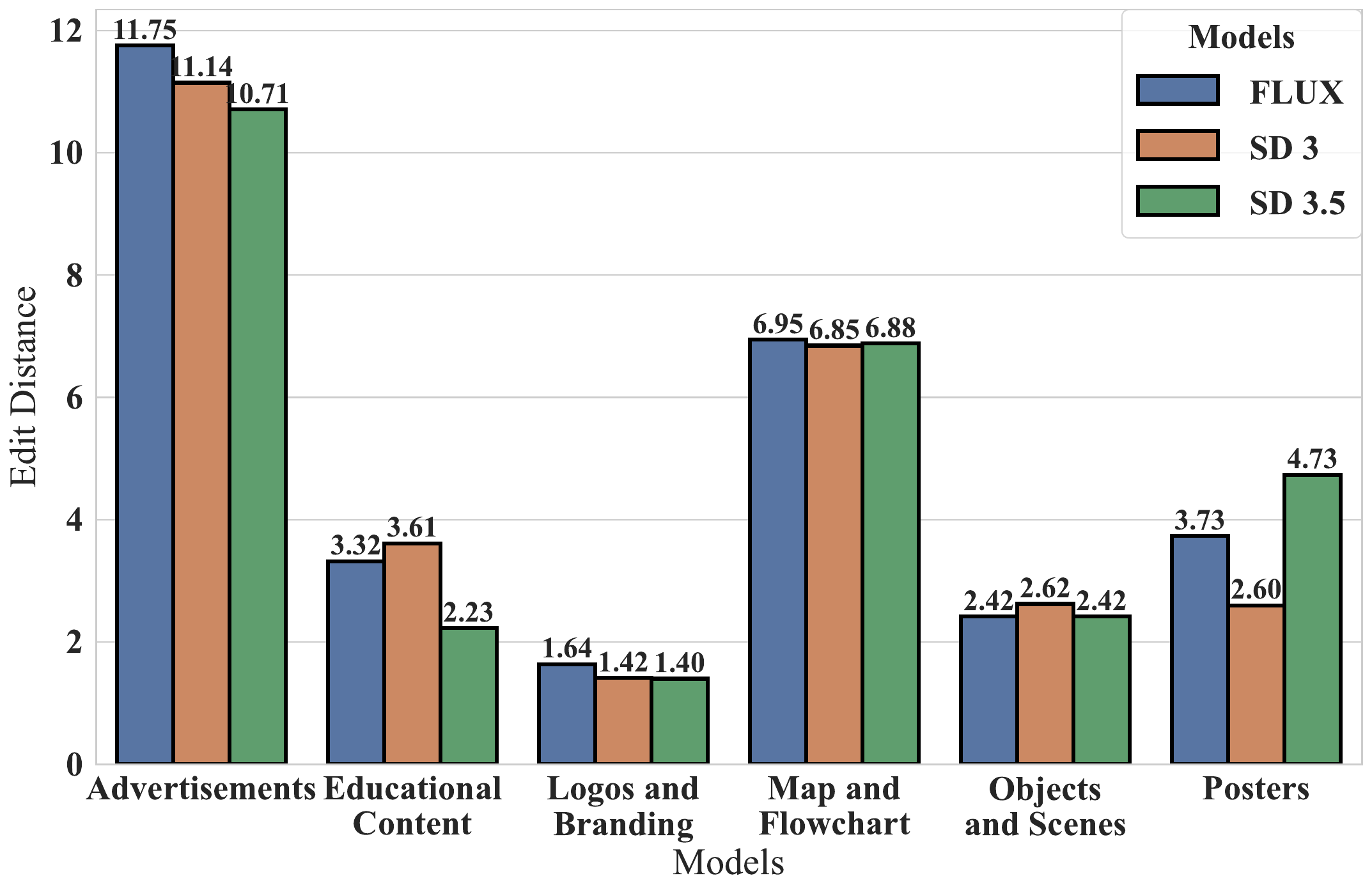}
        \centering
        \caption{Average edit distance using real-world prompts.}
        \label{fig:ed_vs_app}
        \vspace{-0.4cm}
\end{figure}

We initiated our evaluation by utilizing real-world scenario prompts from six main categories (see Figure \ref{fig:ed_vs_app}) and their respective subcategories. For detailed information, please refer to Appendix A.

To assess model performance, we analyzed the average edit distance of generated texts, employing OCR models (see Appendix B for details on the OCR and algorithms used). Categories with more diverse and complex prompts, such as advertisements,  exhibit higher average edit distances compared to other categories like logos and branding.

The poor performance observed in the three SOTA T2I models prompted us to conduct a more granular analysis. We divided our analysis into two main areas: text and prompt. This involved testing single words, phrases, and long texts, alongside two sets of simple and complex prompts, to individually assess the models and identify their specific failure points. The subsequent sections delve deeper into these findings.

\subsection{Impact of text \& prompt complexity}
We benchmark seven SOTA T2I models: Flux \cite{flux}, SD 3 \cite{sd3}, SD 3.5 \cite{sd3}, SD-XL \cite{SDXL}, DeepFloyd \cite{DeepFloydIF}, AnyText \cite{tuo2023anytext}, and Glyph Control \cite{GlyphControl}. We assess the models' capabilities in generating images with accurately integrated text from prompts. To this end, we randomly selected texts and prompts, covering various evaluation tasks. Preliminary tests with other models, such as SD 2.1 \cite{Hierarchical}, Pixart-$\alpha$ \cite{pixart}, and TextDiffuser \cite{Textdiffuser}, yielded unsatisfactory results, and these were excluded from further analysis. The selected models demonstrated decent performance in text embedding, setting a clear benchmark for future use.

\begin{figure*}[t]
\centering
    % \begin{minipage}[b]{0.45\textwidth}
        \centering
        \includegraphics[width=\textwidth]{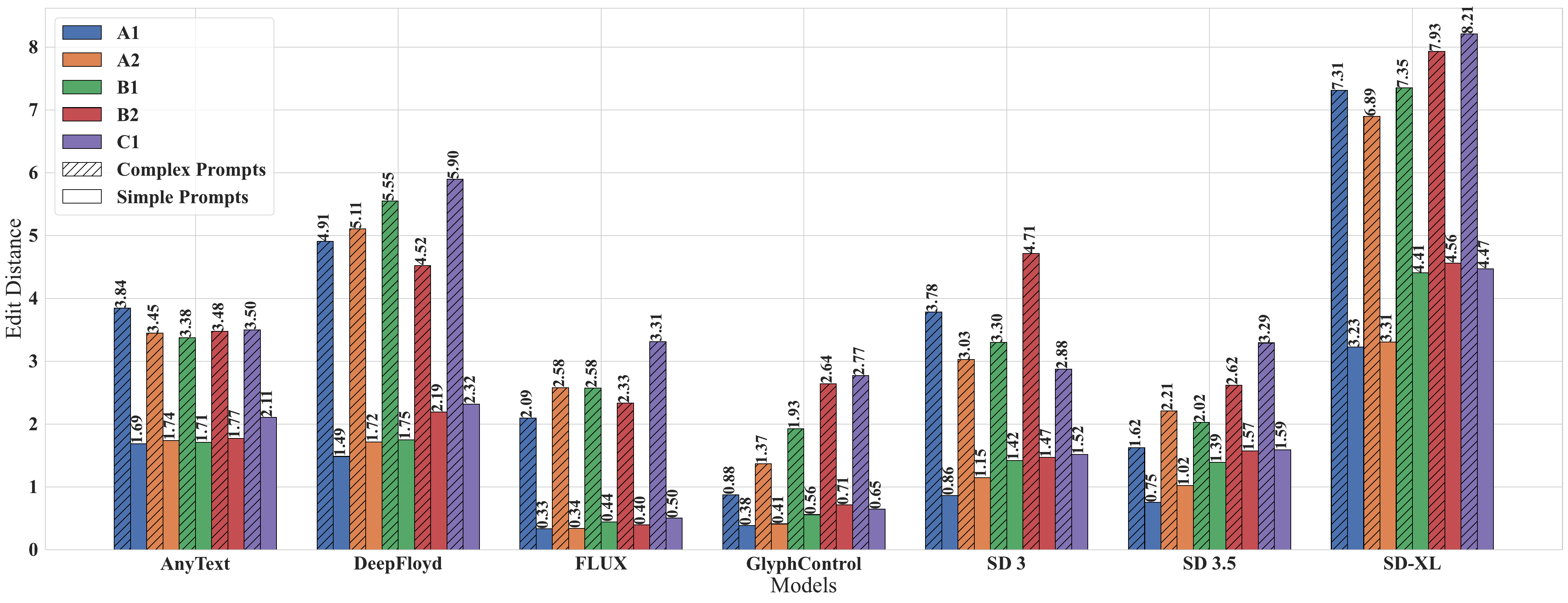}
        \centering
        \caption{Average edit distance using different types of prompts and single words, compared across multiple models.}
        \label{fig:category}
\end{figure*}

\paragraph{Word/text complexity}
Using the Oxford Dictionary's word difficulty classifications (A1 to C1), we analyzed the average edit distance across different word complexity groups using two sets of simple and complex prompts. The results show that although there is a slight increase in average edit distance from simpler words (A1) to more complex ones (C1), the differences are minimal. The observed pattern (see Figure \ref{fig:category}) follows the sequence C1 > B2 > B1 > A2 > A1, indicating a subtle gradation in complexity. These findings suggest that the performance of the T2I models is not significantly affected by the single word complexity as defined by the Oxford classifications.

\begin{table*}[!ht]
\centering
 \caption{CLIP score analysis among multiple T2I models utilizing  complex prompts and single words.}
\small
\resizebox{0.6\textwidth}{!}{%
  \begin{tabular}{@{}lccccccc@{}}
  \toprule
 Models & AnyText & DeepFloyd & FLUX & GlyphControl & SD 3 & SD 3.5 & SD-XL \\ \midrule
CLIP Score & 0.334 & 0.338 & 0.337 & 0.338 & 0.336 & 0.345 & 0.330 \\
  \bottomrule
  \end{tabular}
}
\vspace{-0.4cm}
\label{tab:clip}
\end{table*}

Building upon this, we extended our analysis to compare simple prompts with complex prompts (with single words) that include multiple descriptors, actions, or contexts. In this comparison, we observed a clear decrease in performance as the level of detail in the prompts increased. The added details in these prompts may introduce ambiguities or conflicting instructions, making it challenging for the model to determine which elements to prioritize in the generated image. Although T2I models are trained on vast datasets, they may still struggle with intricate linguistic structures or less common vocabulary found in complex prompts, leading to misinterpretations. 

To ensure that the models were following the prompts, we tested the CLIP scores over complex prompts. As shown in Table \ref{tab:clip}, there are only slight differences between the CLIP scores of the models, even though they exhibit significant variations in terms of edit distances. This suggests that while the models may capture the general content of the prompts to a similar extent, the precision and quality of the generated images, as reflected by the edit distances, can differ substantially.

Consequently, this performance drop highlights the current limitations of T2I models in handling increased prompt complexity and underscores areas for future improvement in model training and architecture.

\paragraph{Word frequency analysis}
We hypothesized that model performance on text generation might correlate with the frequency of words in the training dataset. One of our hypotheses was to determine whether the model can recall, understand, or remember specific spellings and words from its training dataset. We compiled a list of frequent words from the COCO and LAION caption datasets to test this. We evaluated the models against this list to explore any potential relationship between word frequency and edit distances.

Our analysis revealed a slight correlation between word frequency and edit distances, as shown in Table \ref{tab:cor}. We observed a small negative correlation, indicating that edit distances tend to decrease slightly as word frequency increases. However, the correlation coefficient is small, suggesting this effect is not statistically significant. Thus, while a slight trend indicates better performance for more frequent words, word frequency alone is not a reliable predictor of a model's ability to generate text in images accurately.

\begin{table*}[t]
\centering
\caption{The correlation between edit distance and word frequency across models.}
\small
\resizebox{0.6\textwidth}{!}{%
  \begin{tabular}{@{}lccccccc@{}}
  \toprule
  Models & AnyText & DeepFloyd & FLUX & GlyphControl & SD 3 & SD 3.5 & SD-XL \\ \midrule
  COCO Dataset & -0.04 & -0.07 & -0.01 & -0.05 & -0.09 & -0.08 & -0.05 \\
  LAION Dataset & -0.08 & -0.19 & -0.04 & -0.12 & -0.19 & -0.22 & -0.20 \\
  \bottomrule
  \end{tabular}
}
\vspace{-0.4cm}
\label{tab:cor}
\end{table*}

\begin{figure}[t]
    \centering
    \includegraphics[width=0.5\textwidth]{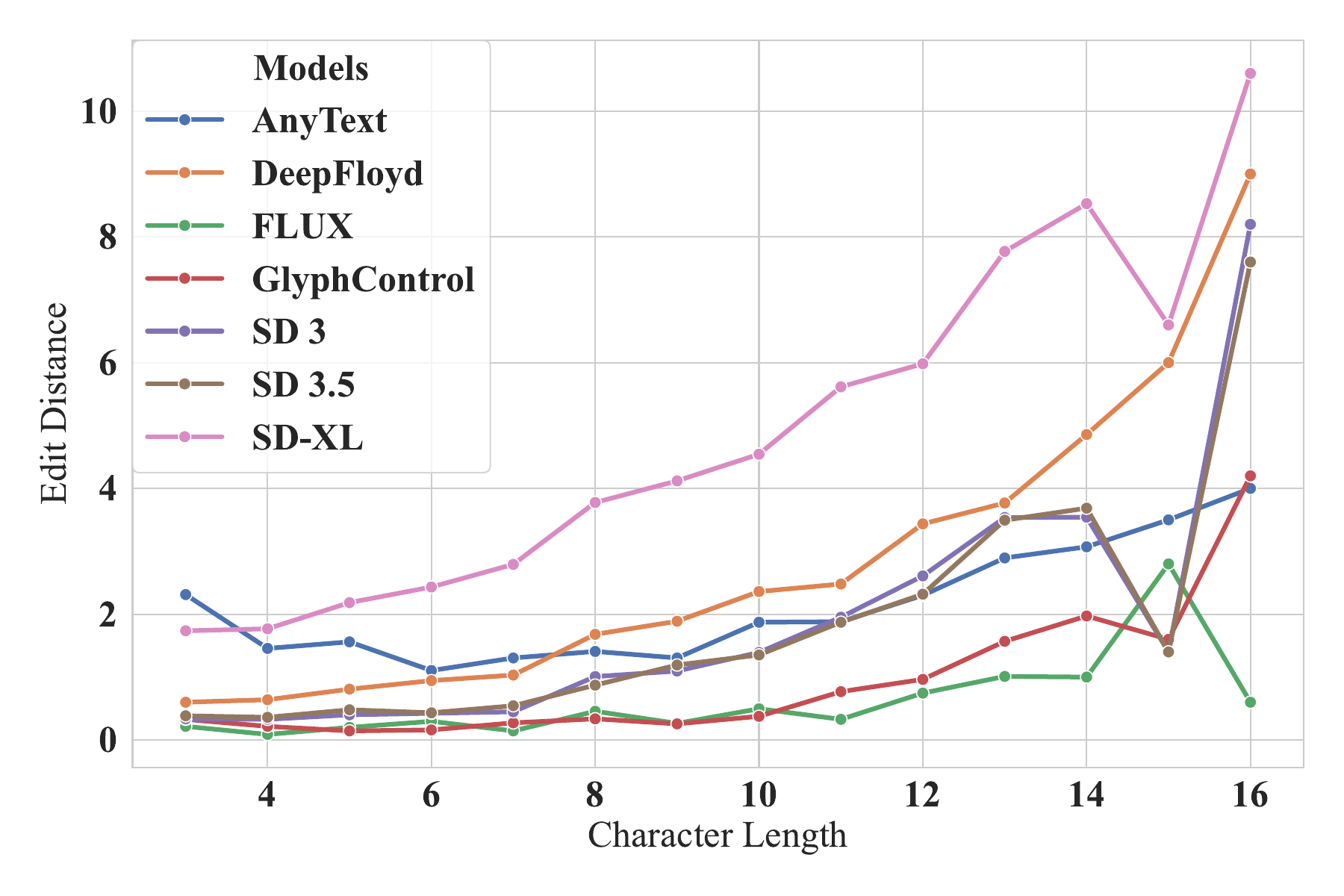}
    \caption{Longer input texts tend to have a higher average edit distance and lead to lower image quality or legibility, further compounding the difficulty in accurately extracting the text.}
    \label{fig:leng}
    % \vspace{-0.4cm}
\end{figure}

\begin{figure}[t]
\centering
        \includegraphics[width=0.5\textwidth]{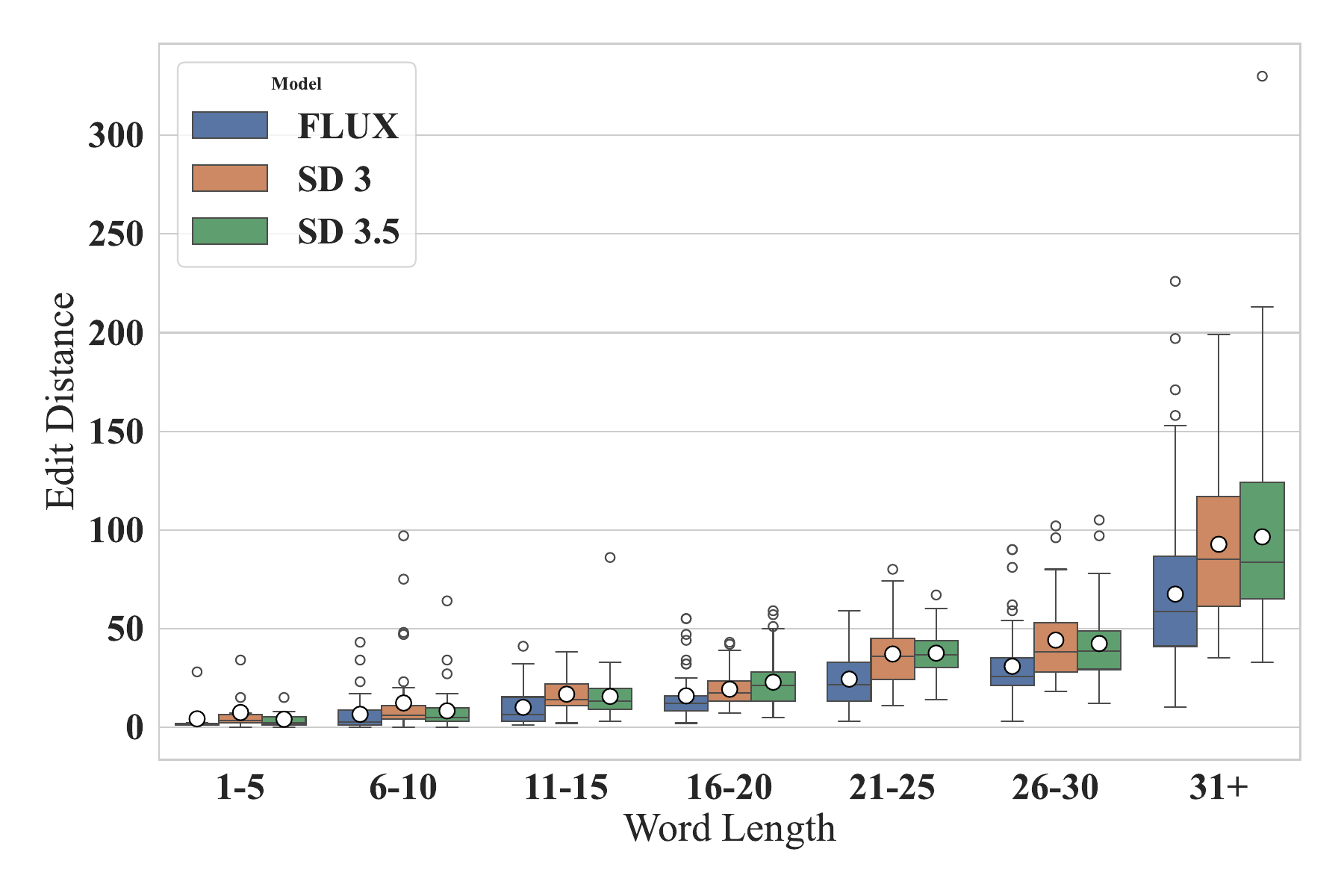}
        \centering
        \caption{Average edit distance of best models across various text lengths, evaluated using simple prompts and long texts.}
        \label{fig:ed_vs_quote}
        \vspace{-0.4cm}
\end{figure}

\paragraph{Word/text length}
Given that text complexity, importance, and frequency do not significantly affect model performance, we focused on word length, analyzing words ranging from 3 to 16 characters (see Figure \ref{fig:leng}) and phrases and sentences spanning 2 to 70 words (see Figure \ref{fig:ed_vs_quote}).

Our results indicate that model performance degrades as word length increases. Interestingly, models performed better with two words totaling 16 characters than with a single 16-character word, suggesting that shorter segments are easier to process. For example, "predict" and "able" are easier to generate accurately than "unpredictable."  Longer words introduce more opportunities for errors, as each additional character increases the likelihood of deviations from the expected text.

Although Figure \ref{fig:leng} shows that Anytext and GlyphControl appear to outperform SD 3 and SD 3.5, this observation is misleading due to limitations in the evaluation metric. Anytext and GlyphControl often fail to generate long text, as a result, the edit distance metric may suggest better performance because shorter or incomplete outputs inherently result in a smaller edit distance, despite lacking semantic accuracy. For the analysis of longer texts, we focused on the three best-performing models, SD 3, SD 3.5, and Flux, that consistently generated coherent long texts to ensure a fair and meaningful evaluation of model performance on longer text generation tasks.

\begin{figure}[]
\centering
        \centering
        \includegraphics[width=0.45\textwidth]{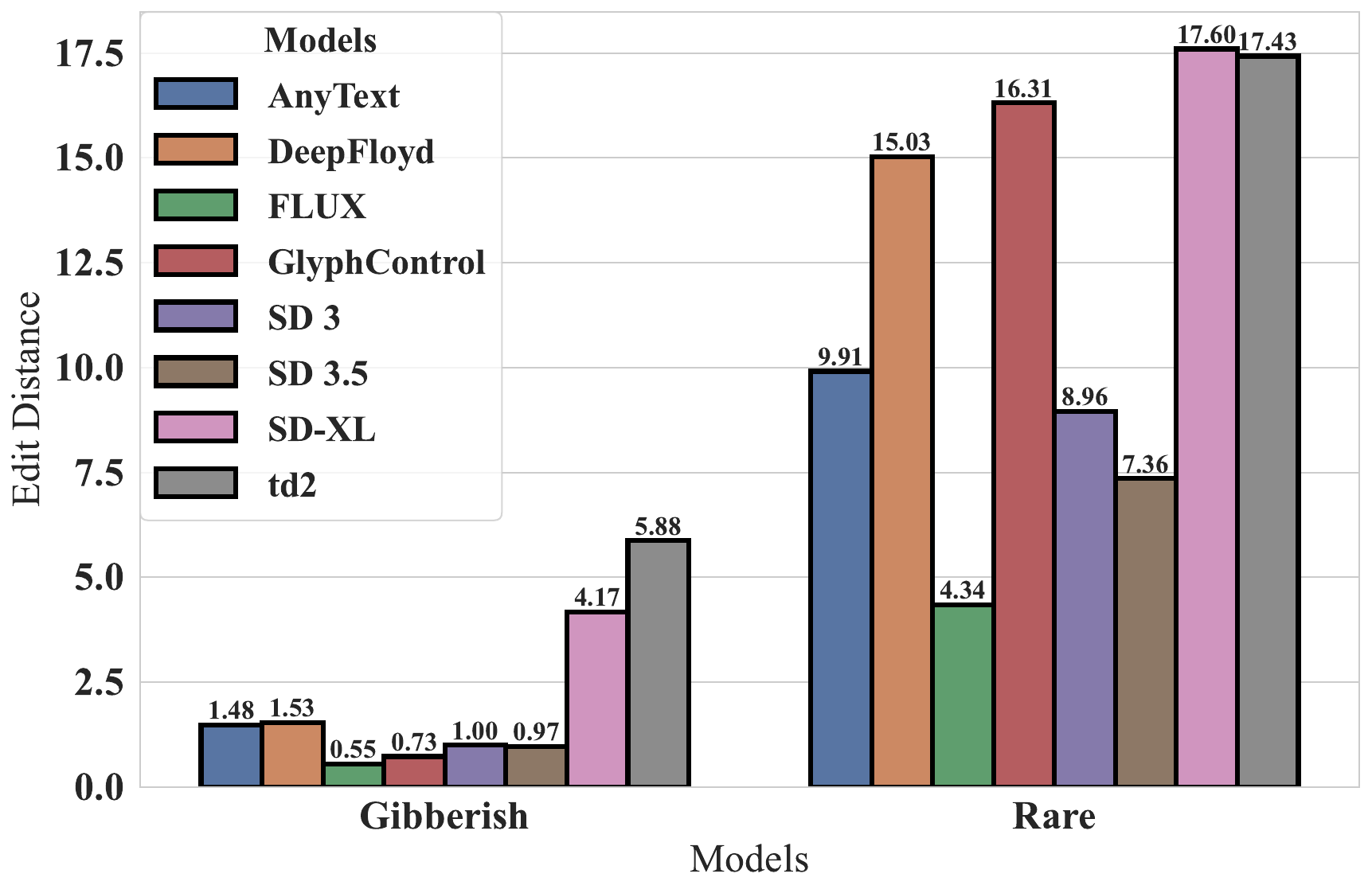}
        \centering
        \caption{Average edit distance for categories rare and gibberish text, compared across multiple models.}
        \label{fig:correct}
        \vspace{-0.4cm}
\end{figure}

\paragraph{Misspelled text, numbers, and special characters}
To evaluate the robustness of T2I models, we crafted prompts that incorporated misspelled words, gibberish text, long rare words, special characters, and numbers. This experiment aimed to assess the models' ability to interpret and depict intended text despite deviations from standard spelling and syntax, thereby gaining insights into whether the models could recognize and correct inaccuracies or generate the text exactly as provided. 
Our results (see Figure \ref{fig:correct}) indicate that rare words pose significant challenges for the models. Although the models recognize these rare words as meaningful and attempt to generate them, they often fail to do so accurately. Conversely, models performed better with gibberish text, possibly because the lack of semantic pressure allows them to focus on visual coherence. These insights highlight the complexity of T2I models and the nuanced challenges models face with different types of textual input, which are crucial for improving training and rendering reliability.

\subsection{Human evaluation}
To complement the automated evaluation using OCR and edit distance, which provided a quantitative measure of the models' ability to generate text within images, we recognized the need for human insight to capture nuances that machines might miss. Therefore, we conducted a comprehensive human evaluation to assess how well the images generated by T2I models followed the prompts and produced accurate, clear text.
We evaluated 1,000 randomly selected images, and each image was assessed by at least 2 human evaluators who answered two questions: whether the image follows the prompt (Prompt Following) and whether the text in the image is accurate and clear (Text Accuracy). 

\begin{figure}[t]
\centering
        \includegraphics[width=0.5\textwidth]{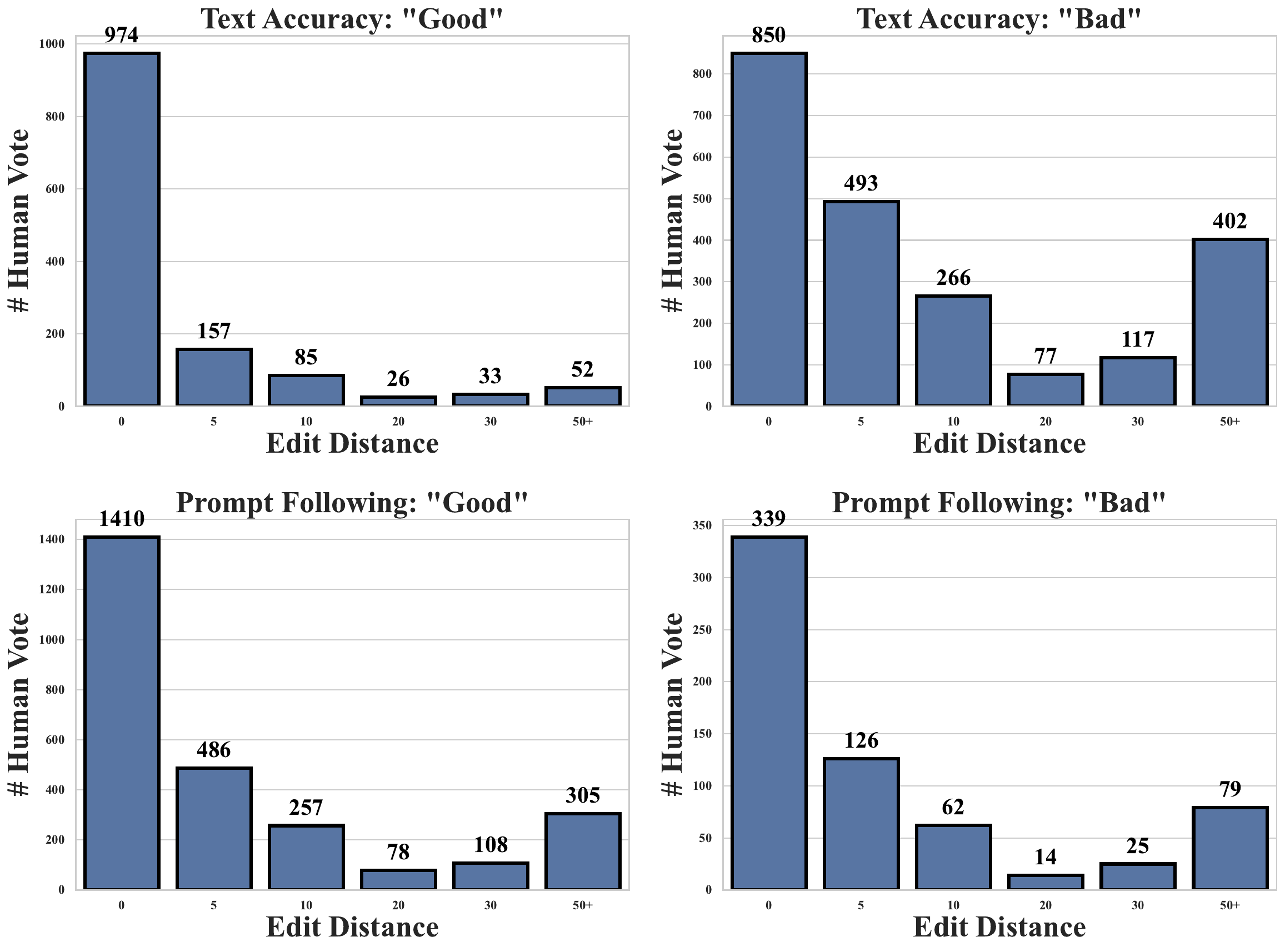}
        \centering
        \caption{Aligning edit distance scores with human assessment standards.}
        \label{fig:hu}
        \vspace{-0.4cm}
\end{figure}

The correlation between edit distance and text accuracy is -0.20, indicating that as the edit distance increases, the accuracy and clarity of the text decrease. The Figure \ref{fig:hu} illustrates that with increment in edit distance, the percentage of "good" counts for text accuracy drops drastically. In contrast, the correlation between edit distance and prompt following is -0.06, suggesting a fairly weak negative trend. This finding corroborates our CLIP scores, implying that while the models can generate images that align with the prompts, they struggle with producing precise and legible text within those images. Further details including inter-rate agreement analysis are provided in Appendix C.

One key factor contributing to these observations is the diverse range of models employed in our study. The reason we observe "Bad" text accuracy when the edit distance is less than 5 and "Good" text accuracy when the edit distance exceeds 50 is the wide range of models used, from Texdiffuser to SD 3.5 and Flux. Some models even fail with single words, while others perform adequately on texts containing up to 70 words.

\subsection{Impact of VAE performance}

\begin{figure*}[!]
    \centering
    \includegraphics[width=\textwidth]{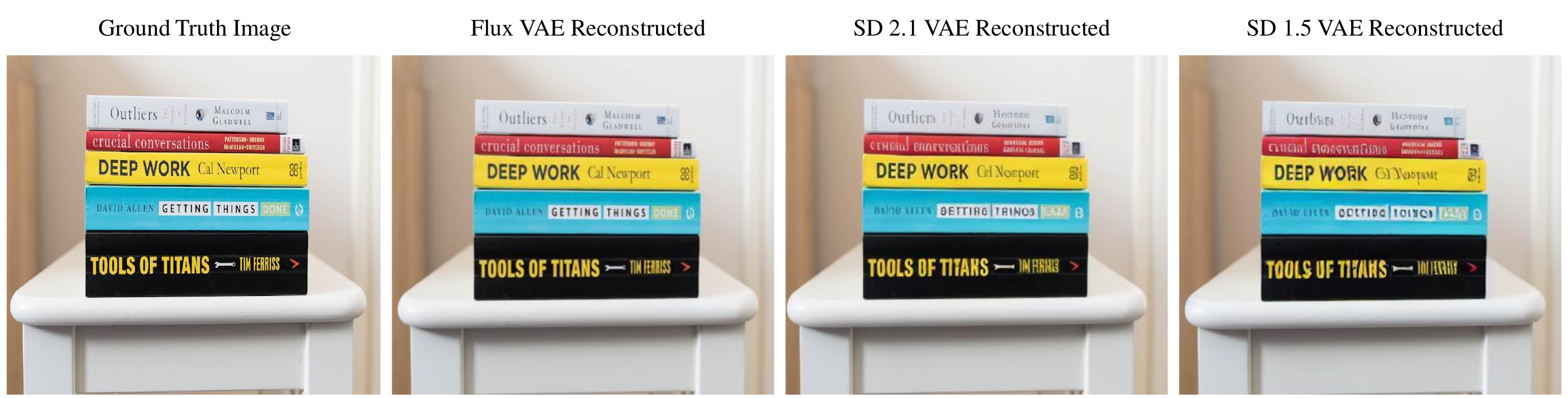}
    \caption{Decreases in quality of VAE reconstructed images leading to non-readable letters. Photo by \href{https://unsplash.com/@jeroendenotter?utm_content=creditCopyText&utm_medium=referral&utm_source=unsplash}{Jeroen den Otter} on \href{https://unsplash.com/photos/several-books-on-top-of-table-inside-room-1SA__aN_I2U?utm_content=creditCopyText&utm_medium=referral&utm_source=unsplash}{Unsplash}   
  }
    \label{fig:vae}
    % \vspace{-0.4cm}
\end{figure*}

Building upon our analysis of how texts and prompts influence the performance of T2I models, it becomes evident that the interplay between textual input and model responsiveness is a critical determinant of successful image generation. Meanwhile, the generation process does not rely solely on prompts and model understanding of texts and prompts. The VAE plays a fundamental role in translating textual embeddings into coherent visual representations, particularly when embedding text within images. Understanding its functionality and limitations is essential to comprehensively evaluate the overall effectiveness of T2I models.

Thus, we hypothesized that the VAE component serves as a critical bottleneck affecting text fidelity in diffusion models. To validate this hypothesis, we conducted a comprehensive investigation into the VAE's role in text reconstruction. This section delineates our methodology, findings, and the profound implications of the VAE's performance on text rendering within generated images.

To thoroughly assess the effectiveness of VAE components in accurately reconstructing textual information, we selected VAEs from the SD 1.5, SD 2.1, and Flux models for evaluation. Our dataset comprised 1,000 images containing textual content sourced from diverse media such as advertisements, web pages, book covers, posters, and billboards. These images were obtained from internet sources (\cite{unsplash, pixabay} and featured a wide range of text characteristics, including varying font sizes, styles, colors, and alignments, all with a minimum resolution of 640 pixels in width.

Each image was encoded and decoded using its respective VAE to produce a reconstructed version. We then employed OCR models (see Appendix B for details on the OCR models) to extract textual content from both the original (ground truth) and the VAE-reconstructed images. By comparing the OCR-extracted texts, we focused on two critical metrics: word retention, which measures the proportion of exactly retained words, and partial accuracy, which assesses the similarity of partially correct words using letter-level edit distance. (See Appendix D for detailed word retention and partial accuracy calculations).

\noindent {\bf VAE is a Bottleneck: }Our analysis revealed significant discrepancies in text retention and reconstruction accuracy among the VAEs. The SD 1.5 model exhibited a mean word retention rate of only 39\%, while the SD 2.1 model demonstrated a modest improvement with a retention rate of 42\%. In contrast, the Flux model achieved a retention rate of 51\%, indicating better performance in maintaining whole words during reconstruction. These results suggest that all models struggle to preserve textual content effectively.

In terms of partial accuracy, the SD 1.5 model had an average letter edit distance of 127, and the SD 2.1 model slightly outperformed it with an average of 122. The Flux model, however, achieved 107. A smaller edit distance indicates a closer approximation to the original text at the character level; nonetheless, all models exhibited substantial deviations from the ground truth (see Figure \ref{fig:vae}).

These findings indicate that the VAE component acts as a significant bottleneck in diffusion models concerning text fidelity. The low word retention rates and high edit distances highlight the inadequacy of current VAEs in encoding and reconstructing textual information accurately. This inability to preserve text leads to considerable degradation in text quality within the reconstructed images.(See Appendix D for more examples).

Consequently, this bottleneck fundamentally limits the effectiveness of T2I models in generating images with clear and accurate text. The current performance of VAEs restricts the potential of diffusion models to produce high-quality images where text accuracy is paramount. Addressing the limitations of the VAE component is imperative for advancing the capabilities of diffusion models in text rendering. Future work should focus on developing more robust mechanisms within the model architecture to faithfully capture and reconstruct textual elements.

\section{Conclusion}
We present TextInVision benchmark, which establishes clear, specific criteria tailored to assess the unique challenges of integrating text into images. This includes evaluating the text's legibility, accuracy, and contextual relevance, ensuring a thorough performance analysis. The TextInVision Benchmark’s prompts are chosen with specific purposes in mind, designed to evaluate the model's performance in realistic and relevant scenarios. Each category and subcategory of prompts is selected to mirror real-world applications. This strategic approach ensures that the evaluation is thorough and meaningful, providing insights into the practical capabilities of the models. While other benchmarks often fail to provide insights into why T2I models struggle with accurate text generation, TextInVision includes analyses highlighting potential bottlenecks in the text rendering process by examining different components, including the VAE. By providing a detailed assessment and pinpointing specific challenges, TextInVision serves as a crucial tool for advancing the field of visual text generation. We hope that this benchmark will guide future research efforts, fostering the development of more accurate and contextually relevant T2I models capable of seamlessly integrating text into images.

\section{Acknowledgments}

MP, AC, CB, and YY are supported by US NSF RI grant \#2132724. 
We thank the NSF NAIRR initiative, the Research Computing (RC) at Arizona State University (ASU), and \href{https://www.cr8dl.ai/}{cr8dl.ai} for their generous support in providing computing resources.
The views and opinions of the authors expressed herein do not necessarily state or reflect those of the funding agencies and employers.

{
    \small
    \bibliographystyle{ieeenat_fullname}
    \bibliography{main}
}

\clearpage
\setcounter{page}{1}
\maketitlesupplementary

\section*{A. TextInVision prompt set}
\label{sup:A}
In this section, we detail the structure of the TextInVision prompt set. As previously discussed, the TextInVision separates the prompts from the textual content that needs to be embedded within images. This separation allows for independent evaluation of model capabilities and aids in pinpointing the causes of errors. Each prompt is associated with specific attributes to facilitate result selection and analysis. For example:

\begin{quote} \textbf{Prompt:} A paper with the words "Vacation calories don't count. Right? - Unknown" written on it.

\textbf{Attributes:} \begin{itemize} \item \textbf{Prompt Type:} Simple \item \textbf{ID:} s119938 \item \textbf{Text:} "Vacation calories don't count. Right? - Unknown" \item \textbf{Character Length:} 41 \item \textbf{Word Count:} 7 \item \textbf{Oxford Category:} None \item \textbf{Contains Rare Words:} No \item \textbf{Contains Gibberish:} No \item \textbf{Contains Special Characters:} Yes \item \textbf{Contains Numbers:} No \item \textbf{Is a Sentence:} Yes \item \textbf{LAION Frequency:} 0 \item \textbf{COCO Frequency:} 0 \item \textbf{Text Source:} \url{https://www.shutterfly.com/ideas/what-to-write-in-a-holiday-card/} \end{itemize} \end{quote}

Another example is:

\begin{quote} \textbf{Prompt:} A science fair with students presenting projects, a display board titled "explore" with diagrams and data, judges evaluating.

\textbf{Attributes:} \begin{itemize} \item \textbf{Prompt Type:} Complex \item \textbf{ID:} c119237 \item \textbf{Text:} "explore" \item \textbf{Character Length:} 7 \item \textbf{Word Count:} 1 \item \textbf{Oxford Category:} B1 \item \textbf{Contains Rare Words:} No \item \textbf{Contains Gibberish:} No \item \textbf{Contains Special Characters:} No \item \textbf{Contains Numbers:} No \item \textbf{Is a Sentence:} No \item \textbf{LAION Frequency:} 467 \item \textbf{COCO Frequency:} 8 \item \textbf{Text Source:} \url{https://www.oxfordlearnersdictionaries.com/wordlists/oxford3000-5000} \end{itemize} \end{quote}

In these examples, the \emph{Text Source} attribute denotes the origin of the textual content. The detailed attributes provide insights into various aspects of the text, such as length, complexity, and frequency in existing datasets, which are crucial for analyzing model performance under different conditions.

\begin{figure*}[]
    \centering
    \includegraphics[width=0.8\textwidth]{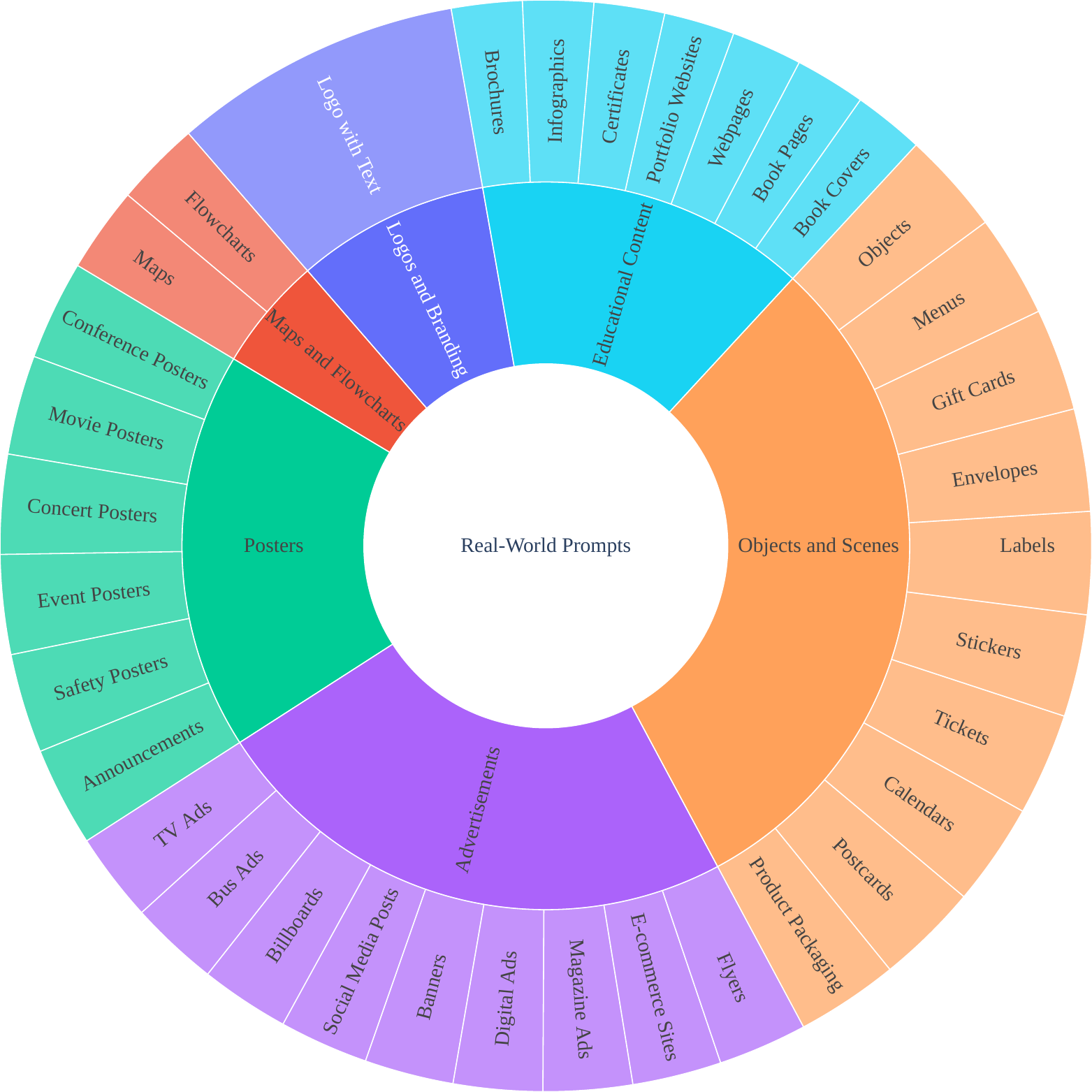}
    \caption{Breakdown of the TextInVision dataset's real-world prompts into six primary categories and their subcategories, highlighting the diversity and proportional representation within the dataset}
    \label{fig:realcat}
    % \vspace{-0.4cm}
\end{figure*}

In addition to general and complex prompts that include various words, phrases, and longer texts, we have incorporated prompts derived from real-world situations that often require images containing visual text. By focusing on real-world scenarios, we aim to enable meaningful evaluations and push the boundaries of current models toward practical applications across different domains.

An example from the \emph{Advertisements} category is:

\begin{quote} \textbf{Prompt:} A local fair with signs at all entrances reading the text "Free Entry Today".

\textbf{Attributes:} \begin{itemize} \item \textbf{Group:} Advertisements \item \textbf{Text:} Free Entry Today \item \textbf{Character Length:} 16 \item \textbf{Word Count:} 3 \item \textbf{ID:} r112277 \end{itemize} \end{quote}

To provide a clearer picture of these real-world prompts, Figure~\ref{fig:realcat} illustrates their distribution across six primary categories, each further divided into subcategories that highlight their respective contributions to the overall dataset.

\begin{figure*}[]
    \centering
    \includegraphics[width=\textwidth]{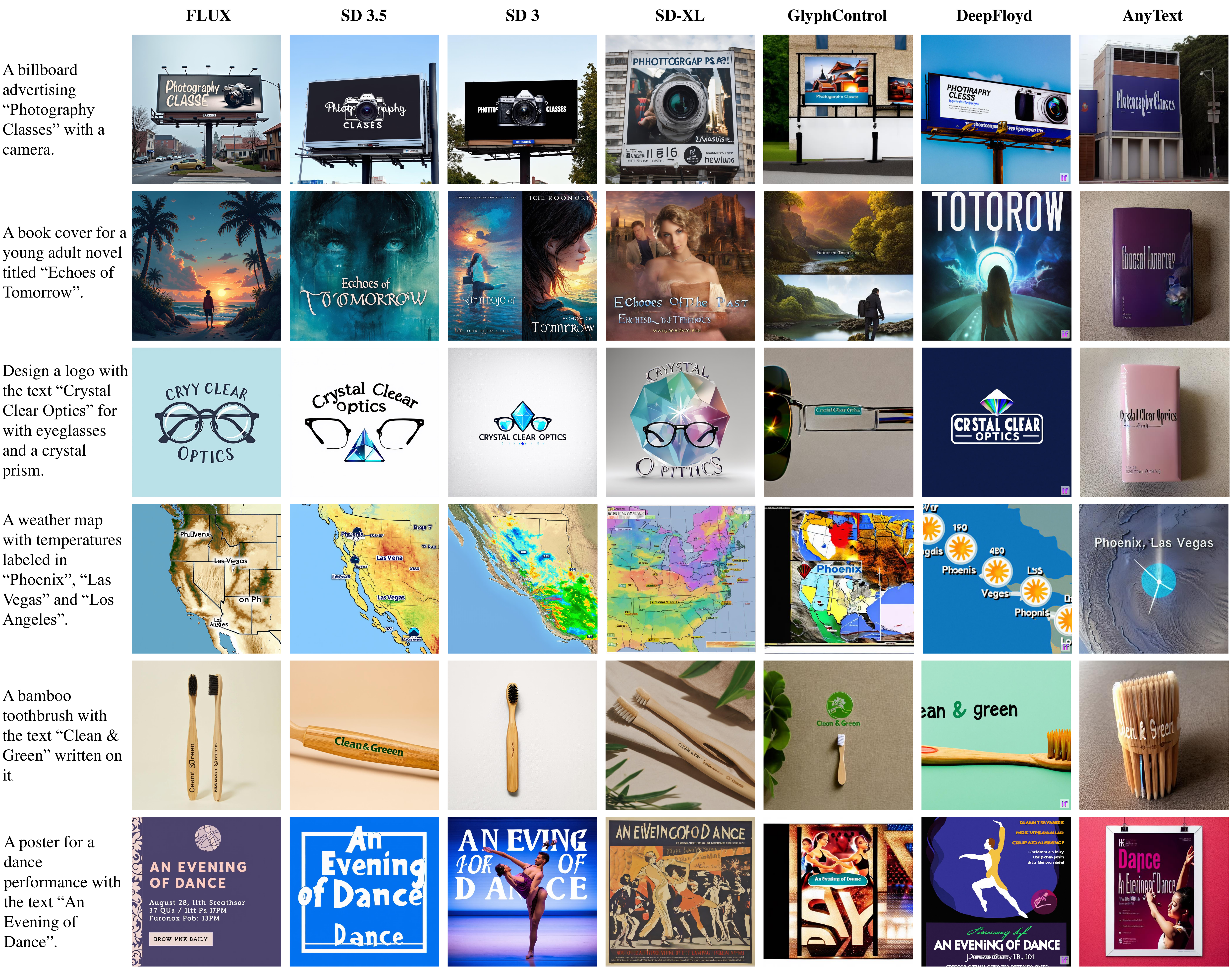}
    \caption{Examples of real-world prompts and the images generated by seven models, demonstrating their performance.   
}
    \label{fig:sampe_su}
    % \vspace{-0.4cm}
\end{figure*}

To showcase how the models perform on prompts from different categories, we present more examples in Figure \ref{fig:sampe_su}. This figure includes prompts and the corresponding images generated by the seven models evaluated in this study. These examples illustrate the models' capabilities and limitations when handling various types of textual content within images.

\section*{B. OCR \& edit distance score}

To extract text from images, we initially utilized three OCR models: Llava Next \cite{liu2024llavanext}, EasyOCR \cite{easyocr2020}, and PaddleOCR \cite{paddle}. As our evaluation progressed, we opted to use EasyOCR for the remaining tasks due to its faster processing speed and consistent performance. While Llava Next was capable of accurate text recognition, it lacked sufficient reliability for our requirements. PaddleOCR did not offer significant advantages over EasyOCR, reinforcing our decision to rely exclusively on EasyOCR.

To comprehensively analyze the OCR results, we employed three methods, each providing a unique perspective on the accuracy and quality of the text extraction:

\begin{itemize} 

\item \textbf{Edit Distance} \cite{edit}: Measures the minimum number of single-character edits (insertions, deletions, or substitutions) required to transform one string into another. This metric is useful for evaluating the degree of similarity between the OCR output and the ground truth, especially in cases of partial matches. 

\item \textbf{Longest Common Subsequence (LCS)}: Identifies the longest sequence of characters that appear in both the OCR result and the ground truth in the same order, but not necessarily consecutively. This method allows for gaps or misspellings and provides insight into the overall similarity between the two texts. 

\item \textbf{Longest Ordered Match (LOM)}: Focuses on finding the longest sequence of characters that appear in both texts while maintaining their original order. This metric offers insight into how well the OCR preserves the text structure and sequence. 

\end{itemize}

For example, when comparing the ground truth "Digital Dreamscapes" with the OCR result "Dlgitoi Draseampes":
\begin{itemize}
    \item Edit Distance
    \begin{itemize}
            \item Edits needed:
        \begin{enumerate}
            \item Substitute 'i' with 'l'
            \item Substitute 'a' with 'o'
            \item Substitute 'l' with 'i'
            \item Substitute 'e' with 'a'
            \item Substitute 'a' with 's'
            \item Substitute 'm' with 'p'
            \item Substitute 'c' with 'e'
            \item Substitute 'e' with 's'
            \item Delete the final 's'
        \end{enumerate}
    \item Total edits: 12 (A lower number signifies a closer match between the OCR result and the ground truth.)
    \end{itemize}
\item LCS
\begin{itemize}
    \item LCS: "Dgit Dreamapes"
    \item Length of LCS: 13 
\end{itemize}
\item LOM
\begin{itemize}
    \item LOM: "git eam"
    \item Length of LOM: 6 
\end{itemize}
\end{itemize}

Given that T2I models can produce more text than desired, ensuring that the extracted text is accurate and aligns with the prompts is crucial. Therefore, we used Algorithm \ref{alg:modified_ocr} that incorporates these three methods to analyze the OCR results and evaluate the model's performance. This algorithm systematically evaluates the OCR results by calculating the edit distance and identifying the LCS, providing a comprehensive assessment of the OCR model's performance. 

\begin{algorithm}[]
\caption{Edit distance based score}
\label{alg:modified_ocr}
\begin{algorithmic}[1]
% \Require \textit{ocr\_text}, \textit{exp\_text}
% \Ensure \textit{dist}
\State \textbf{Input:} \textit{ocr\_text}, \textit{exp\_text}
\State \textbf{Output:} \textit{dist}

\State Read \textit{ocr\_text}
\If{\textit{exp\_text} is a substring of \textit{ocr\_text}}
    \State \textit{dist} $\gets$ 0
\Else
    \If{\textit{exp\_text} is a single word}
        \State \textit{words} $\gets$ \textsc{SplitWords}(\textit{ocr\_text})
        \State \textit{max\_lcs\_len} $\gets$ 0
        \State \textit{best\_match} $\gets$ empty string
        \ForAll{\textit{w} in \textit{words}}
            \State \textit{lcs} $\gets$ \textsc{LCS}(\textit{w}, \textit{exp\_text})
            \If{\textsc{Length}(\textit{lcs}) $>$ \textit{max\_lcs\_len}}
                \State \textit{max\_lcs\_len} $\gets$ \textsc{Length}(\textit{lcs})
                \State \textit{best\_match} $\gets$ \textit{w}
            \EndIf
        \EndFor
        \State \textit{dist} $\gets$ \textsc{EditDistance}(\textit{best\_match}, \textit{exp\_text})
    \Else
        \State \textit{exp\_words} $\gets$ \textsc{SplitWords}(\textit{exp\_text})
        \State \textit{ocr\_words} $\gets$ \textsc{SplitWords}(\textit{ocr\_text})
        \ForAll{\textit{w} in \textit{exp\_words}}
            \If{\textit{w} is in \textit{ocr\_words}}
                \State Remove \textit{w} from \textit{exp\_words}
                \State Remove \textit{w} from \textit{ocr\_words}
            \EndIf
        \EndFor
        \State \textit{rem\_exp} $\gets$ \textsc{JoinWords}(\textit{exp\_words})
        \State \textit{rem\_ocr} $\gets$ \textsc{JoinWords}(\textit{ocr\_words})
        \State \textit{dist} $\gets$ \textsc{EditDistance}(\textit{rem\_ocr}, \textit{rem\_exp})
    \EndIf
\EndIf
\State \Return \textit{dist}
\end{algorithmic}
\end{algorithm}

In this algorithm, the variables are \textit{ocr\_text}, representing the OCR output text; \textit{exp\_text}, the expected text to compare against; \textit{dist}, the computed edit distance; \textit{words}, a list of words extracted from \textit{ocr\_text}; \textit{max\_lcs\_len}, the maximum length of the LCS found; \textit{best\_match}, the word from \textit{ocr\_text} that best matches \textit{exp\_text}; \textit{exp\_words} and \textit{ocr\_words}, lists of words from \textit{exp\_text} and \textit{ocr\_text}, respectively; and \textit{rem\_exp} and \textit{rem\_ocr}, the remaining texts after removing common words. The functions used are \textsc{SplitWords(text)}, which splits the text into a list of words; \textsc{JoinWords(words)}, which joins a list of words into a single string; \textsc{LCS(a, b)}, which computes the longest common subsequence between strings \( a \) and \( b \); \textsc{EditDistance(a, b)}, which calculates the edit distance between two strings; and \textsc{Length(s)}, which returns the length of the string \( s \).

\section*{C. Human evaluation}

\begin{figure*}[]
    \centering
    \includegraphics[width=\textwidth]{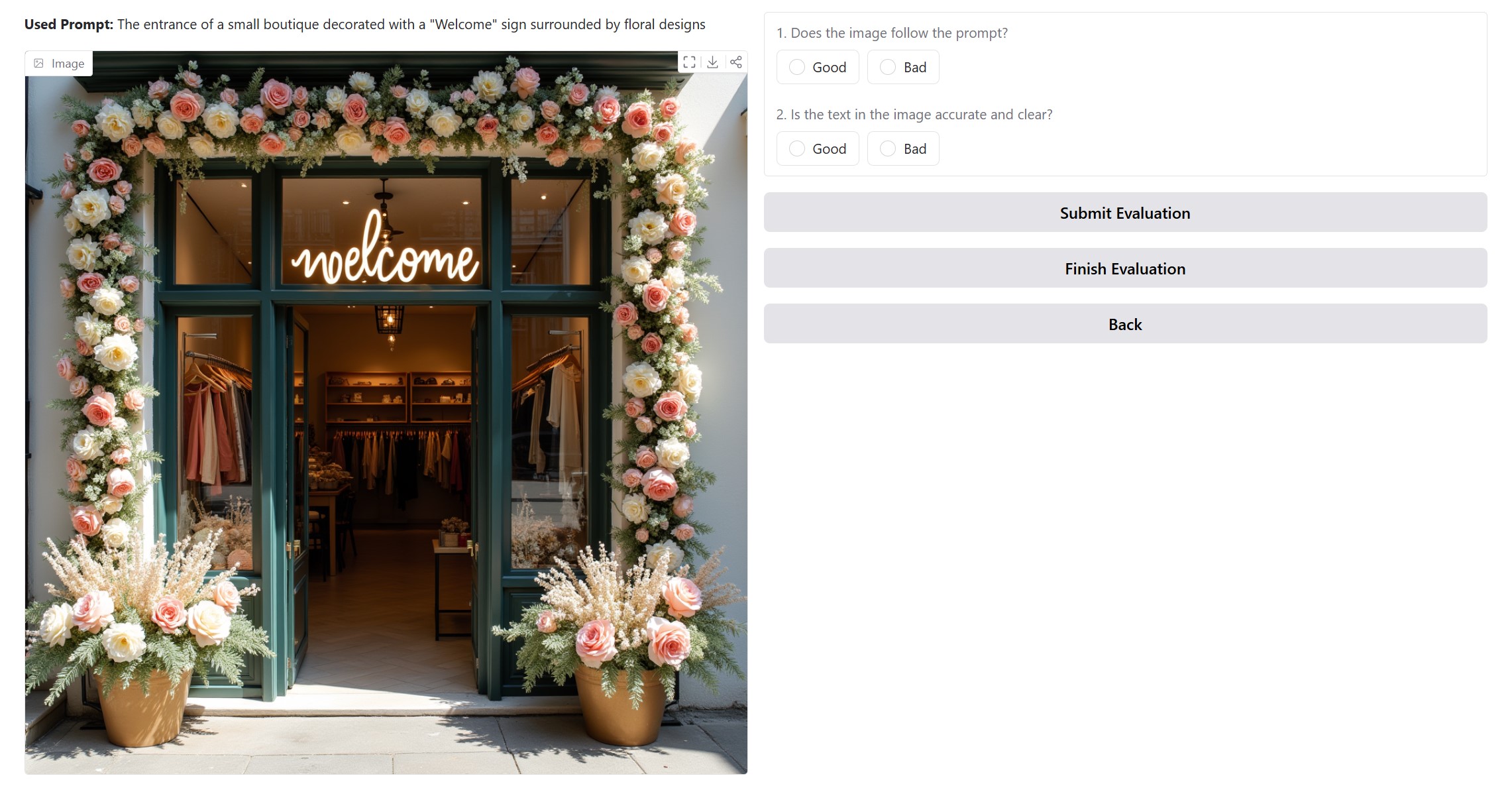}
    \caption{Examples of real-world prompts and the images generated by seven models, demonstrating their performance.   
}
    \label{fig:huggingface}
    % \vspace{-0.4cm}
\end{figure*}

We conducted a human evaluation study to validate our TextInVision prompt set for assessing T2I models' ability to generate images with accurate embedded text, based on the hypothesis that current models struggle to fully reflect prompts in this aspect. The primary objective was to determine whether our edit distance-based score aligns with human judgments and confirms our comparisons of model performance, highlighting their shortcomings.

To achieve this, a total of 66 participants %, all students,at the School of Computing and Artificial Intelligence (SCAI)%,
were recruited for this evaluation, hosted on a Hugging Face Space (see Figure \ref{fig:huggingface}). This accessible and user-friendly platform enabled participants to review images and submit their responses efficiently.

The evaluation dataset consisted of 1,000 images, randomly selected to ensure fairness and comprehensive coverage across all categories and models in our study. We ensured that at least 10 images from each group (combination of category and model) were included to maintain equitable representation. This stratified random sampling approach aimed to minimize selection bias and provide a balanced assessment of each model's performance.

Given the large number of images, it was not feasible for a single person to evaluate all and on average, each participant evaluated 30 images. To assess the consistency, we calculated the agreement per image and the average agreement across all images and participants. For the prompt following criterion, the average agreement was 90.06\%, and for the accuracy criterion, the average agreement was 88.53\%.

These high agreement rates demonstrate that, despite each image being evaluated by different subsets of participants, there was a strong overall consistency in the assessments. This consistency enhances the credibility of our human evaluation methodology and supports the reliability of our conclusions regarding the performance of the T2I models. It indicates that the participants, even when evaluating different images, applied the evaluation criteria in a similar manner, reinforcing the validity of our findings about the models' shortcomings in generating images with accurate embedded text.

\section*{D. Detailed evaluation of VAE}
In this section, we provide an additional information on the evaluation of the VAE component applied to the TexInVision image set. The following subsections offer further insights into the dataset organization and the metrics used in our analysis.

\subsection*{TexInVision image set}

\begin{figure*}[]
    \centering
    
    \begin{subfigure}[]{\textwidth}
        \centering
        \includegraphics[width=\textwidth]{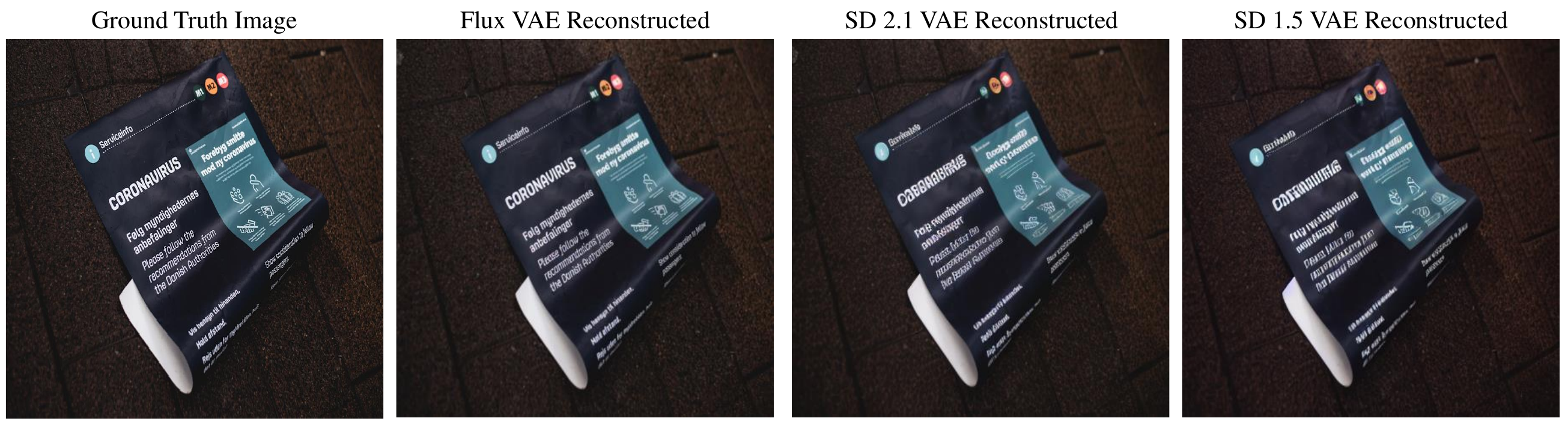}
        \caption{Photo by \href{https://unsplash.com/@mors_dreng}{Hjalte Gregersen} on \href{https://images.unsplash.com/photo-1588324524886-7b9475d2a7d0?crop=entropy&cs=tinysrgb&fit=max&fm=jpg&ixid=M3w2MDM1NDN8MHwxfHNlYXJjaHw0ODB8fGFkdmVydGlzZW1lbnQlMjBwb3N0ZXJ8ZW58MHx8fHwxNzMwMTQwMTc4fDA&ixlib=rb-4.0.3&q=80&w=1080}{Unsplash} \cite{unsplash}.}
        \label{fig:vae1}
    \end{subfigure}
    
    \vspace{0.5cm} % Optional: Adjust the vertical space between subfigures
    
    \begin{subfigure}[]{\textwidth}
        \centering
        \includegraphics[width=\textwidth]{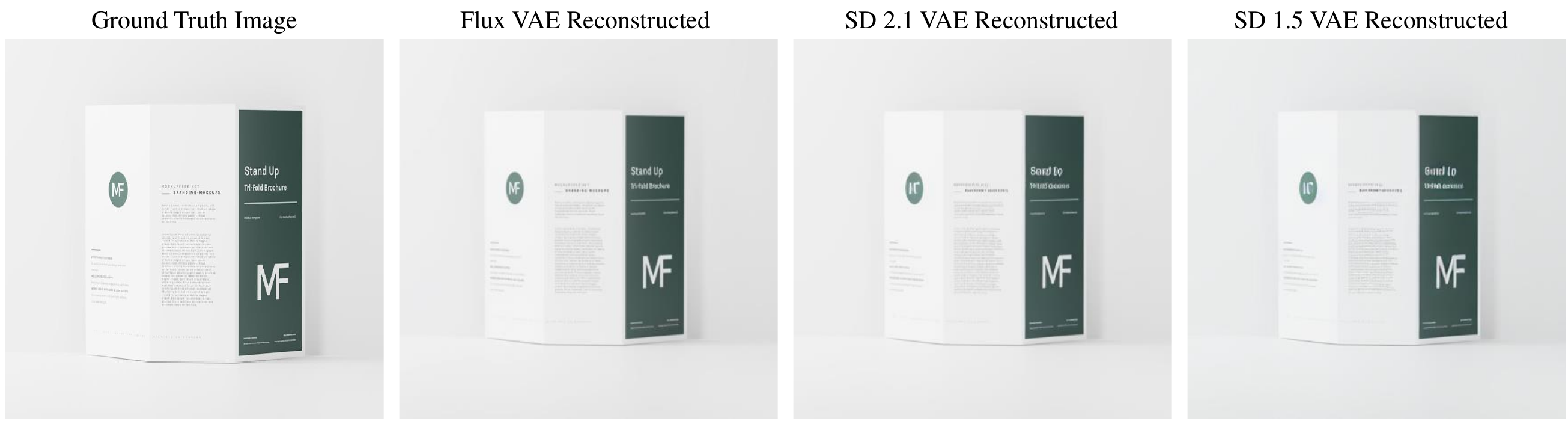}
        \caption{Photo by \href{https://unsplash.com/@mockupfreenet}{Mockup Free} on \href{https://images.unsplash.com/photo-1695634281463-4788ac6ddfdf?crop=entropy&cs=tinysrgb&fit=max&fm=jpg&ixid=M3w2MDM1NDN8MHwxfHNlYXJjaHw0OTF8fGFkdmVydGlzZW1lbnQlMjBwb3N0ZXJ8ZW58MHx8fHwxNzMwMTQwMTkxfDA&ixlib=rb-4.0.3&q=80&w=1080}{Unsplash} \cite{unsplash}.}
        \label{fig:vae2}
    \end{subfigure}
    
    \caption{Assessing VAEs for reconstructing text-embedded images reveals noticeable detail degradation and errors in text representation. }
    \label{fig:vae_combined}
\end{figure*}

The dataset is organized within a JSON file, which facilitates easy access and management of the images. Each entry in the JSON file contains essential metadata, including the source, photographer profile link, and a link to get the image. This structured format ensures efficient handling and retrieval of data for further processing. Figure \ref{fig:vae_combined} presents more examples of VAE reconstructed images from the TextInVision image set.

\subsection*{Word retention \& partial accuracy calculations}

Word Retention (WR) and Partial Accuracy (PA) are essential metrics for comprehensively analyzing OCR results in the evaluation of VAEs. WR measures the proportion of words that are exactly retained by the OCR process compared to the ground truth text. It provides a straightforward metric to assess the accuracy of word-level recognition. 

Let $N_{\text{correct}}$ be the number of words exactly matched between the OCR output and the ground truth and $N_{\text{total}}$ be the total number of words in the ground truth text. Then, $WR$ is calculated as:

\begin{equation}
    WR = \frac{N_{\text{correct}}}{N_{\text{total}}} \times 100\%
    \label{eq:word_retention}
\end{equation}

However, solely relying on exact word matches can overlook subtle yet significant errors within recognized words. This is where PA becomes crucial, as it assesses the similarity of partially correct words by evaluating the letter-level edit distance between the OCR output and the ground truth. 

Let $W_i$ be the $i$-th word in the ground truth,  $\hat{W}_i$ be the corresponding OCR-recognized word,  $d(W_i, \hat{W}_i)$ be the Levenshtein edit distance between $W_i$ and $\hat{W}_i$ and $L(W_i)$ be the number of letters in $W_i$.
Then, $PA$ is calculated as:

\begin{equation}
    PA = \left(1 - \frac{\sum_{i=1}^{N_{\text{total}}} d(W_i, \hat{W}_i)}{\sum_{i=1}^{N_{\text{total}}} L(W_i)}\right) \times 100\%
    \label{eq:partial_accuracy}
\end{equation}

Together, these metrics enable a nuanced evaluation of the VAE's performance in processing and reconstructing textual data, highlighting both its strengths and areas for improvement in maintaining the integrity of the original text.

% {
%     \small
%     \bibliographystyle{ieeenat_fullname}
%     \bibliography{main}
% }
\end{document}